\documentclass[letterpaper]{article} 
\usepackage{aaai25}  
\usepackage{times}  
\usepackage{helvet}  
\usepackage{courier}  
\usepackage[hyphens]{url}  
\usepackage{graphicx} 
\usepackage{natbib}  
\usepackage{caption} 
\usepackage{bm}
\usepackage{multirow}
\usepackage{amsmath,amssymb,amsfonts}

\usepackage{booktabs}
\usepackage{subfigure}
\usepackage{enumitem}

\usepackage{algorithm}
\usepackage{algorithmic}
\usepackage{xcolor}

\usepackage{newfloat}
\usepackage{listings}
\DeclareCaptionStyle{ruled}{labelfont=normalfont,labelsep=colon,strut=off} 
\floatstyle{ruled}
\newfloat{listing}{tb}{lst}{}
\floatname{listing}{Listing}
\def\method{ABPEM}

\title{Attention Bootstrapping for Multi-Modal Test-Time Adaptation}
\author{
    Yusheng Zhao\textsuperscript{\rm 1},
    Junyu Luo\textsuperscript{\rm 1},
    Xiao Luo\textsuperscript{\rm 2,*},
    Jinsheng Huang\textsuperscript{\rm 1}, \\
    Jingyang Yuan\textsuperscript{\rm 1},
    Zhiping Xiao\textsuperscript{\rm 3,*},
    Ming Zhang\textsuperscript{\rm 1,}\thanks{Corresponding authors.}
}
\affiliations{
    \textsuperscript{\rm 1}State Key Laboratory for Multimedia Information Processing, \\School of Computer Science, PKU-Anker LLM Lab, Peking University, Beijing, China\\
    \textsuperscript{\rm 2}Department of Computer Science, University of California, Los Angeles, CA, USA  \\
    \textsuperscript{\rm 3}Paul G. Allen School of Computer Science and Engineering, University of Washington, Seattle, WA, USA\\
    \{yusheng.zhao, luojunyu, hjs\}@stu.pku.edu.cn, xiaoluo@cs.ucla.edu, \\\{yuanjy, mzhang\_cs\}@pku.edu.cn, patxiao@uw.edu

}

\usepackage{bibentry}

\begin{document}

\maketitle

\begin{abstract}
Test-time adaptation aims to adapt a well-trained model to potential distribution shifts at test time using only unlabeled test data, without access to the original training data. While previous efforts mainly focus on a single modality, test-time distribution shift in the multi-modal setting is more complex and calls for new solutions. This paper tackles the problem of multi-modal test-time adaptation by proposing a novel method named Attention Bootstrapping with Principal Entropy Minimization (\method{}). We observe that test-time distribution shift causes misalignment across modalities, leading to a large gap between intra-modality discrepancies (measured by self-attention) and inter-modality discrepancies (measured by cross-attention). We name this the \emph{attention gap}. This attention gap widens with more severe distribution shifts, hindering effective modality fusion. To mitigate this attention gap and encourage better modality fusion, we propose \emph{attention bootstrapping} that promotes cross-attention with the guidance of self-attention. Moreover, to reduce the gradient noise in the commonly-used entropy minimization, we adopt \textit{principal entropy minimization}, a refinement of entropy minimization that reduces gradient noise by focusing on the principal parts of entropy, excluding less reliable gradient information. Extensive experiments on the benchmarks validate the effectiveness of the proposed \method{} in comparison with competing baselines.
\end{abstract}

\section{Introduction}
Multi-modal learning \cite{blikstein2013multimodal, xu2023multimodal} has recently attracted increasing attention, with a wide range of applications in many fields, including autonomous driving \cite{zheng2023autofed}, video understanding \cite{lee2023lecture}, sentiment analysis \cite{yu2021learning}, and robotics \cite{krauhausen2024bio}. Recent advances in this field often encode each modality into tokens and utilize transformers to learn the embedding \cite{yao2020multimodal, zhang2022transformer}. Then, the modalities are often fused with the attention mechanism. Although this paradigm has achieved promising results, it assumes that the test data have the same distribution as the training data, which may fail to hold in the wild \cite{niu2023towards, tang2023distribution, liang2024comprehensive}.

\begin{figure}
    \centering
    \includegraphics[width=\linewidth]{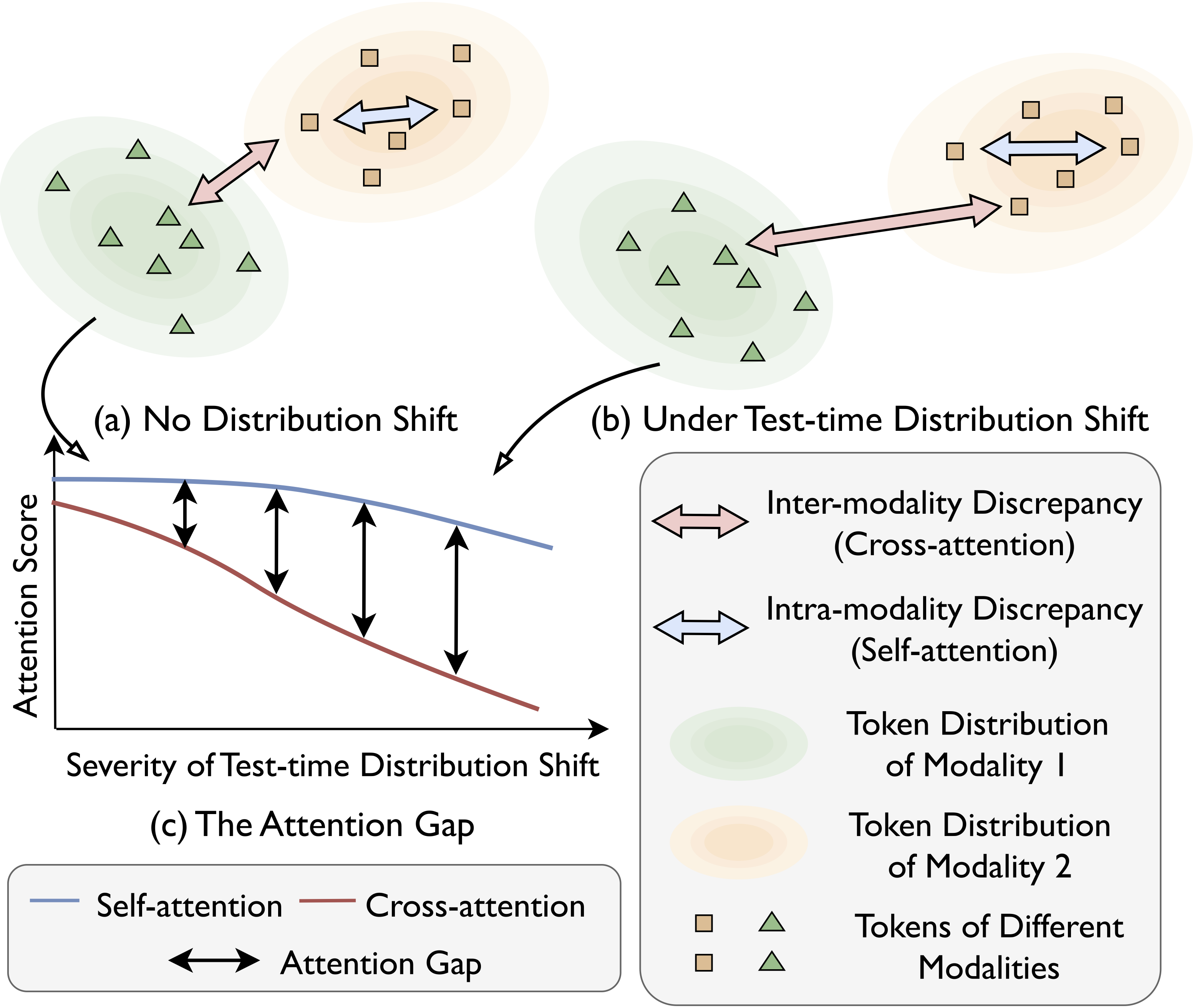}
    \vspace{-3mm}
    \caption{During test time, the distribution shift typically has a larger impact on the inter-modality discrepancy than intra-modality discrepancy, leading to an increasing attention gap.}
    \vspace{-4mm}
    \label{fig:motivation}
\end{figure}

To tackle the challenge of test-time distribution shift, test-time adaptation has emerged as a promising solution as it assumes neither the labels of test data (which is more practical) nor the access of training data (which protects privacy). Recently, many test-time adaptation methods have been proposed \cite{wang2020tent, boudiaf2022parameter, chen2022contrastive, nguyen2023tipi, karmanov2024efficient}. However, most existing test-time adaptation methods focus on the uni-modal setting. In practice, multi-modal test-time adaptation is more challenging. As is shown in Figure \ref{fig:motivation}, the test-time distribution shift causes not only intra-modality changes (blue arrows) but also inter-modality changes (red arrows), and the latter can potentially undermine the model's ability to effectively align and fuse different modalities as the cross-attention tends to decrease under distribution shift.

Towards this end, we propose a novel method named Attention Bootstrapping with Principal Entropy Minimization (\method{}). \footnote{In this paper, the term bootstrap is used in its idiomatic sense rather than the statistical sense.} As illustrated in Figure \ref{fig:motivation}, when the attention-based model is challenged by test-time distribution shift, intra-modality discrepancies increase mildly, while inter-modality discrepancies increase significantly. This is indicated by a mild decrease in the raw self-attention score (before softmax) and a sharp decrease in the raw cross-attention score, which leads to the attention gap. Decreased cross-attention hinders the alignment and fusion across modalities, leading to potentially inferior test-time performance. 
To encourage cross-attention and decrease inter-modality discrepancies, a naive approach is to minimize the difference between modalities. However, such a method could cause mode collapse and information loss. 

To solve this problem, we propose attention bootstrapping that promotes cross-attention scores using self-attention scores. Concretely, we model the distribution of raw cross-/self-attention scores and use the distribution of self-attention scores as the anchor to align the distribution of cross-attention scores. This is better than the naive approach, as it also takes into account the inherent reliability of each modality. It is conceivable that when the intra-modality discrepancies become larger under distribution shift (as indicated by low self-attention scores), this modality becomes less reliable. In such cases, we will have a low anchor (low self-attention scores), and the cross-attention scores have a low target, which results in less attention to this modality.

Moreover, to reduce the noise in the self-supervising signals in multi-modal test-time adaptation, we propose principal entropy minimization. Entropy minimization \cite{wang2020tent, zhang2023domainadaptor, gao2024unified} is a commonly used technique in test-time adaptation in the absence of ground truth labels. However, the computation of entropy involves the model's predictions on every class, reliable ones and unreliable ones. Under test-time distribution shift, the model's predictions on the less-likely classes (classes with lower probabilities) become less reliable, and the gradients of them become noisy. Therefore, the proposed principal entropy minimization excludes the less-likely classes and focuses only on the more-likely (principal) classes, which reduce gradient noise.

Extensive experiments on the benchmarks demonstrate the effectiveness of the proposed method. The contribution of this work is summarized as follows:
\begin{itemize}
    \item We tackle the problem of multi-modal test-time adaptation, which is practical yet under-explored, and propose a novel method named Attention Bootstrapping with Principal Entropy Minimization (\method{}).
    \item We reveal that test-time distribution shift causes modality misalignment, and propose attention bootstrapping to encourage modality alignment and fusion.
    \item We propose principal entropy minimization that focuses on the principal part of the entropy and reduces the gradient noise in traditional entropy minimization.
\end{itemize}
\section{Related Works}

\textbf{Test-time Adaptation.} Test-time adaptation tackles the problem of distribution shift during test-time, but it assumes neither the knowledge of test data labels nor the access of training data. It is a practical setting since test data labels are hard to obtain, and it also protects privacy \cite{zhu2021federated, tan2023federated}. Recently, a number of methods have been proposed to solve this problem, utilizing entropy minimization \cite{wang2020tent, kundu2020universal, liu2021source, mummadi2021test, lee2023towards, lee2024entropy}, sample selection \cite{litrico2023guiding, pei2023uncertainty}, normalization layer tuning \cite{hu2021mixnorm, yang2022test, lim2023ttn, wu2024test}, representation invariance \cite{nguyen2023tipi, wang2023feature, chen2023improved, ma2024invariant}, self-supervised learning \cite{liu2021ttt++, azimi2022self, ma2024improved}, and generative methods \cite{gao2023back, prabhudesai2024test, tsai2024gda}. While they have achieved remarkable performance, existing efforts mainly focus on a single modality. In multi-modal setting, test-time distribution shift causes not only intra-modal discrepancies but also inter-modal discrepancies. 
There are some works on multi-modal test-time adaptation, but this work differs from them. 
Shin et al. \cite{shin2022mm} focus on the specific task of 2D-3D joint segmentation, whereas our method is designed for more general multi-modal settings. 
Yang et al. \cite{yang2024test} reveal the challenge of reliability bias caused by multi-modal distribution shifts, and propose READ to tackle the reliability problem. By comparison, we observe a different phenomenon named the attention gap that hinders modality fusion, and propose \method{} to promote modality fusion under distribution shifts.

\smallskip
\noindent\textbf{Multi-modal Learning.} Learning from multi-modal data is an essential topic of deep learning \cite{he2021transrefer3d, yang2022learning, zhao2022target, ma2024cross, huang2024mmevalpro}. Recently, there are increasing attention in alignment and fusion of different modalities \cite{prakash2021multi, xu2023murf}. Efforts have been devoted to ensure effective fusion in adverse settings, including modality imbalance \cite{zhou2020improving, peng2022balanced, fan2023pmr}, missing modality \cite{ma2021smil, ma2022multimodal, woo2023towards, wang2024gradient}, and distribution shift \cite{liu2023osan, tang2024source, xia2024achieving}. However, these works focus mainly on the model's training stage, and in resource-limited scenarios, tuning the model's backbone might be infeasible. Moreover, these algorithms often require the labels of the data, which is hard to obtain in practice. This work differs from existing studies, as it explores adapting the model online and during test-time, which is a more practical setting under limited computation resources.

\begin{figure*}
    \centering
    \includegraphics[width=\linewidth]{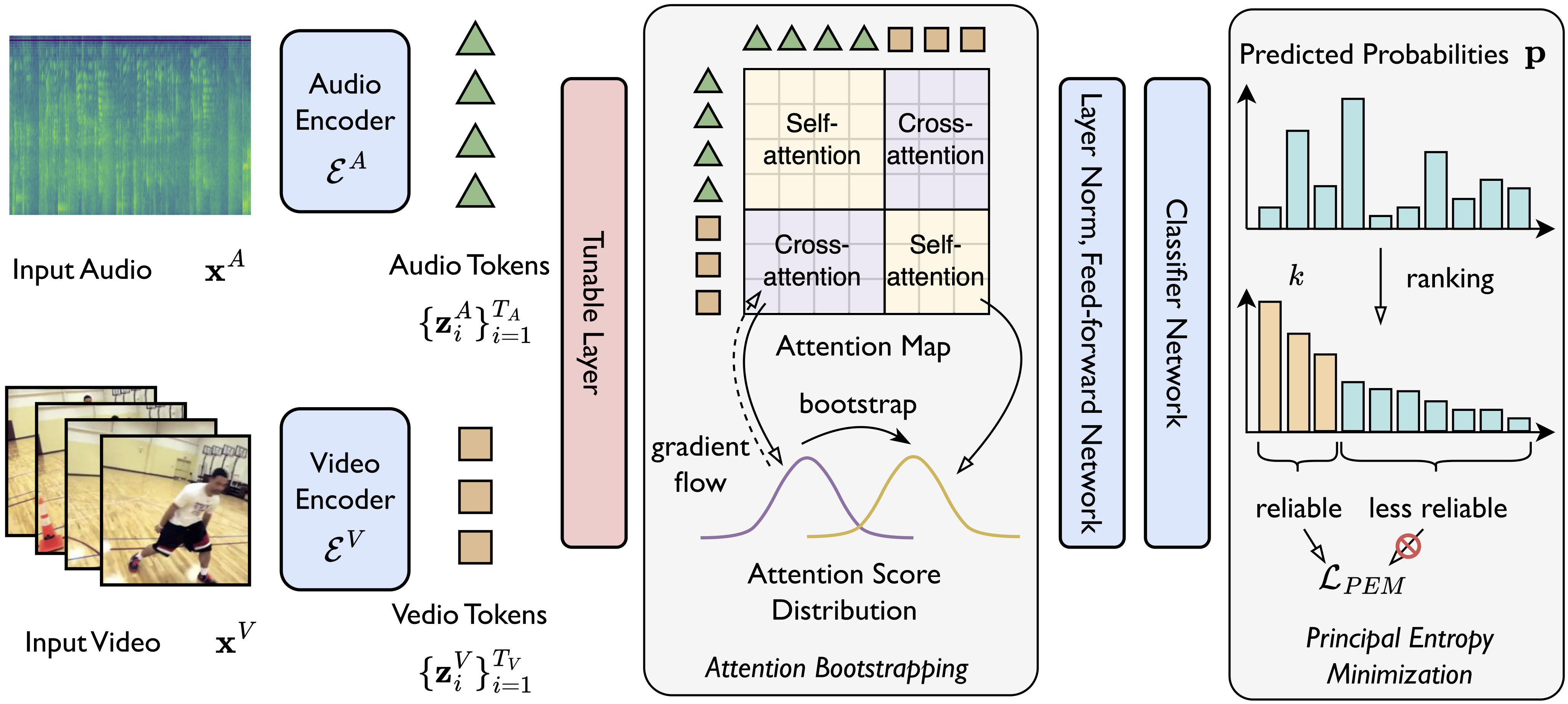}
    \caption{The framework of the proposed \method{}.}
    \vspace{-3mm}
    \label{fig:framework}
\end{figure*}
\section{Methodology}
\subsection{Problem Definition} 
For simplicity, and without loss of generality, we use two modalities (audio, denoted as $A$, and video, denoted as $V$) to present the algorithm. The input of each modality is denoted as $\bm x^A$ and $\bm x^V$. The multi-modal learning system encodes the inputs into two sets of tokens in the hidden space, \emph{i.e.} $\{ \bm z_i^A \}_{i=1}^{T_A}$ and $\{\bm z_i^V \}_{i=1}^{T_V}$ (where $T_A$ and $T_V$ are the numbers of tokens), using modality-specific encoders, \emph{i.e.} $\mathcal E^A$ and $\mathcal E^V$. Then, an attention-based fusion module $\mathcal F$ is used that combines the two sets of tokens and outputs the probability distribution, \emph{i.e.} $\bm p=\mathcal F(\{ \bm z_i^A \}_{i=1}^{T_A}, \{\bm z_i^V \}_{i=1}^{T_V})$. The probability distribution is denoted as $\bm p = [p_1, \cdots, p_C] \in \Delta^{C-1}$, where $C$ is the number of classes, and $\Delta^{C-1}$ is the probability simplex. In multi-modal test-time adaptation, the model $\mathcal M=(\mathcal E^A, \mathcal E^V, \mathcal F)$ has already been trained on the training set $\mathcal D_{tr}=\{(\bm x_i^A, \bm x_i^V, y_i)\} _{i=1} ^{N_{tr}}$, where $N_{tr}$ is the size of the training set and $y_i$ is the label. However, the task does not assume access of $D_{tr}$, and instead, the goal is to improve the model's performance using unlabeled test data $\mathcal D_{te}=\{(\bm x_i^A, \bm x_i^V)\} _{i=1} ^{N_{te}}$, where $N_{te}$ is the size of the test set. For practicability, we fix $\mathcal E^A$, $\mathcal E^V$ and only tune a small part of the parameters in $\mathcal F$.

\subsection{Framework Overview} 
The framework of the proposed \method{} is illustrated in Figure \ref{fig:framework}. During test time, the input audio and video are encoded by $\mathcal E^A$ and $\mathcal E^B$ to obtain hidden space representations (\emph{i.e.} tokens). Then, the tokens from different modalities are concatenated and a tunable layer is applied to generate queries, keys and values for each token, which are used for attention. The attention bootstrapping is used on the attention map that aligns the distributions of cross- and self-attention. Subsequently, the attended tokens are processed by layer normalization, the feed-forward network, and the classifier network to generate predicted probabilities $\bm p\in \Delta^{C-1}$. Finally, principal entropy minimization takes top $k$ reliable classes and computes the principal part of the entropy, which is part of the objective.

\subsection{Attention Bootstrapping}
The attention mechanism \cite{vaswani2017attention} is the most widely used paradigm for modality fusion \cite{nagrani2021attention, zong2023mcomet}. However, when the model is challenged by multi-modal test-time distribution shift, inter-modality discrepancy experiences more increase than intra-modality discrepancy, which leads to a widening gap between self-attention and cross-attention. Attention bootstrapping aims to bootstrap cross-attention using self-attention. Under the aforementioned paradigm, the tokens learned by the modality-specific encoders are first concatenated as $\bm Z=[\bm z^A_1, \cdots, \bm z^A_{T_A}, \bm z^V_1, \cdots, \bm z^V_{T_V}]^T$. Subsequently, query, key and value matrices ($\bm Q,\bm K, \bm V$) are computed as:
\begin{equation}\label{eq:qkv}
\begin{aligned}
\bm Q=\bm Z \bm W_Q + &\bm B_Q, \quad \bm K=\bm Z \bm W_K +\bm B_K, \\ \bm V &=\bm Z\bm W_V +\bm B_V,
\end{aligned}
\end{equation}
where $\bm W_{Q,K,V}$ and $\bm B_{Q,K,V}$ are learnable parameters during test time. Then, the unnormalized attention can be computed as follows using queries and keys: 
\begin{equation}\label{eq:unnormalized-attention}
\tilde{\bm A} = \bm Q \bm K^T.
\end{equation}
We decompose $\tilde{\bm A}$ into four parts, \emph{i.e.}
\begin{equation}
\tilde{\bm A} = 
\begin{bmatrix}
\tilde{\bm A}^{A2A} & \tilde{\bm A}^{A2V} \\
\tilde{\bm A}^{V2A} & \tilde{\bm A}^{V2V}
\end{bmatrix},
\end{equation}
where the $X2Y$ superscript denotes the unnormalized attention when modality $X$ is the query and modality $Y$ is the key.
The normalized version of attention scores are used to fuse different modalities:
\begin{equation}\label{eq:softmax}
\bm A=\operatorname{softmax}(\tilde{\bm A}/\sqrt{d}), \quad \bm Z'=\bm A \bm V,
\end{equation}
where $d$ is the dimension of tokens and $\bm Z'$ is the attended token embeddings. 

When the model experiences test-time distribution shifts, modality mismatch may happen, and this will cause both intra- and inter- modality discrepancy. For example, when the input video is subject to distribution shift, the distribution of token embeddings $\{\bm z_i^V \}_{i=1}^{T_V}$ will have slightly higher variance due to the increased uncertainty (intra-modality discrepancy), and they will also shift away from audio tokens (inter-modality discrepancy). The intra-modality discrepancy is signified by the decreasing value of self-attention scores (\emph{i.e.} $\tilde{\bm A}^{V2V}$), and inter-modality discrepancy is shown by the decreasing of cross-attention (\emph{i.e.} $\tilde{\bm A}^{A2V}$). Note that we use the unnormalized attention scores (the ones before softmax), as they better reflect the distance of distributions. Normalized attention scores are influenced by many factors (\emph{e.g.} the number of tokens). To better describe the attention scores, we model them as Gaussian distributions:
\begin{equation}
\begin{aligned}
&P_{A2A}(a)\sim \mathcal N(\mu_{A2A};\sigma^2_{A2A}), \\ \mu_{A2A}=&\operatorname{avg}(\tilde{\bm A}^{A2A}_{ij}), \quad \sigma_{A2A}^2=\operatorname{var}(\tilde{\bm A}^{A2A}_{ij}),
\end{aligned}
\end{equation}
where $i=1,2,\cdots, T_A$ and $j=1,2,\cdots, T_A$. Similarly, $P_{A2V}(a)$, $P_{V2V}(a)$, and $P_{V2A}(a)$ can be defined. 

\begin{figure}
    \centering
    \includegraphics[width=\linewidth]{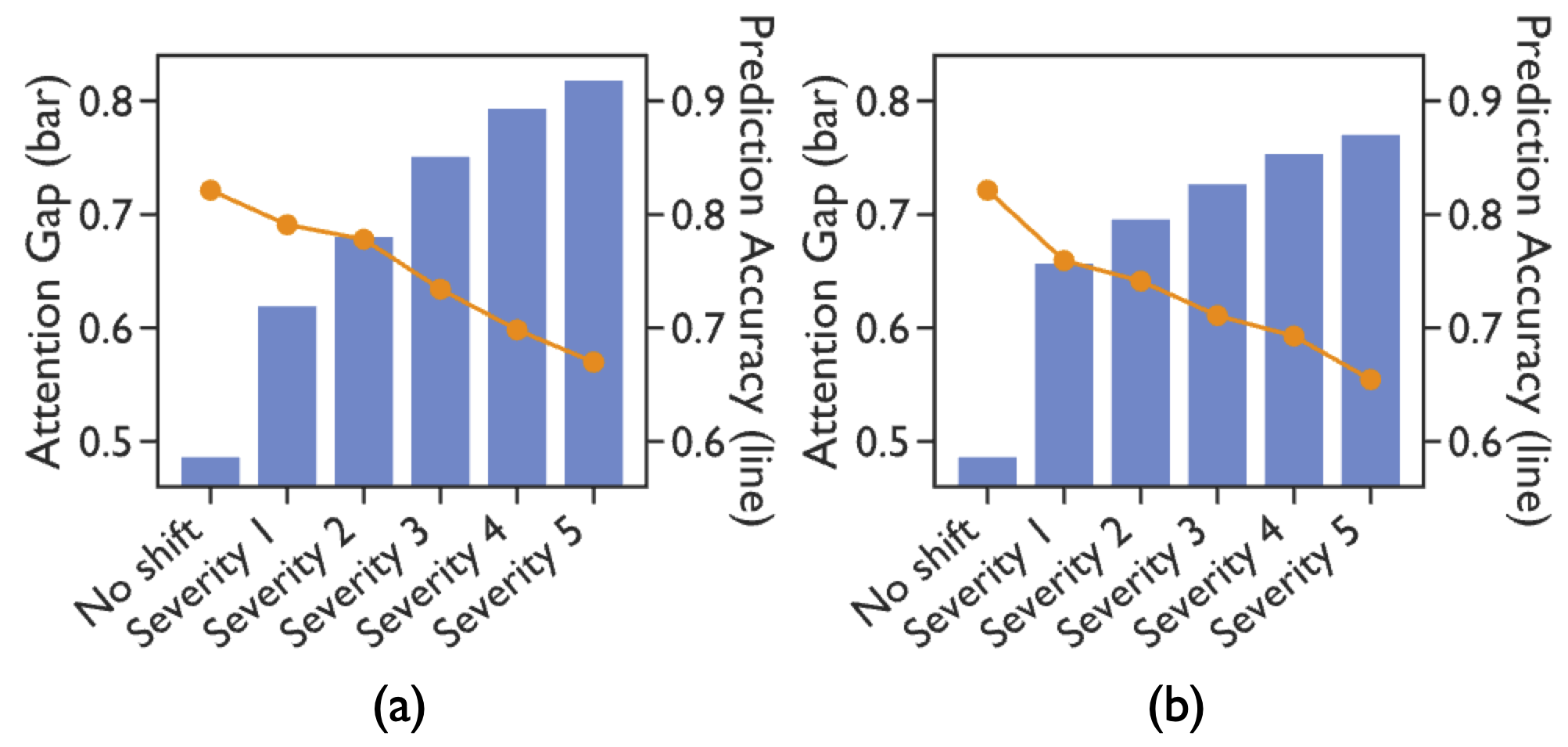}
    \vspace{-3mm}
    \caption{As the test-time distribution shift becomes severer, the attention gap (blue bar plot) tends to increase, and the prediction accuracy (orange line plot) tends to decrease.}
    \label{fig:ab-demo}
    \vspace{-3mm}
\end{figure}

In Figure \ref{fig:ab-demo}, we provide the empirical evidence that the attention gap exists and tends to increase as the test-time distribution shift becomes severer. Specifically, we introduce two types of noise to the vision modality, \emph{i.e.} defocus blur (a) and zoom blur (b), and measure the attention gap  $\mu_{V2V}-\mu_{A2V}$ as well as the prediction accuracy of the model. As can be seen from the figure, when the model faces test-time distribution shift, the attention gap (blue bar plot) generally increases and the prediction accuracy (orange line plot) drops. This suggests that inter-modality dependencies (signified by cross-attention scores, \emph{i.e.} $P_{A2V}(a)$ and $P_{V2A}(a)$) are more affected than intra-modality dependencies (signified by self-attention scores, \emph{i.e.} $P_{A2A}(a)$ and $P_{V2V}(a)$) under distribution shift. Therefore, a feasible solution is to perform attention bootstrapping that uses self-attention scores as anchors to boost cross-attention scores, and thus reduce the attention gap, encouraging modality fusion.

Specifically, we adopt the strategy that minimizes the Kullback-Leibler divergence between the distributions of attention scores, which is formulated as follows:
\begin{equation}
\begin{aligned}
\mathcal D_{KL}(P_{A2V}||P_{V2V}) = \log \frac{\sigma_{V2V}}{\sigma_{A2V}} -\frac{1}{2} \\+ \frac{\sigma_{A2V}^2+(\mu_{A2V}-\mu_{V2V})^2}{2\sigma_{V2V}^2}.
\end{aligned}
\label{eq:kldiv}
\end{equation}
In Eq. \ref{eq:kldiv}, $P_{V2V}$ reflects how the video modality attends itself, and $P_{A2V}$ describes how the audio modality attends the video modality. In other words, $P_{V2V}$ reflects the video's evaluation of itself: $\mu_{V2V}$ is an evaluation of the amount of information relevant to the prediction task, while $\sigma_{V2V}$ is an evaluation of discriminability across tokens.
When the video modality itself has been fully adapted to the distribution shift, it is conceivable that such evaluation is better than the assessment from the audio modality ($\mu_{A2V}$ and $\sigma_{A2V}$) since there are modality discrepancies. 
Thus, our goal is to decrease inter-modality discrepancies so that they are similar to intra-modality ones (not decreased to zero as we want to preserve the natural discrepancies of different tokens).
Therefore, it is reasonable to use $P_{V2V}$ as an anchor to bootstrap $P_{A2V}$. We stop the gradient flow of the self-attention scores $\mu_{V2V}$ and $\sigma_{V2V}$ to avoid influencing the anchor.

Similarly, $\mathcal D_{KL}(P_{V2A}||P_{A2A})$ can be calculated, and the loss objective of attention bootstrapping is written as:
\begin{equation}\label{eq:loss-ab}
\mathcal L_{AB} = \mathcal D_{KL}(P_{A2V}||P_{V2V}) + \mathcal D_{KL}(P_{V2A}||P_{A2A}).
\end{equation}

\begin{figure}
    \centering
    \includegraphics[width=\linewidth]{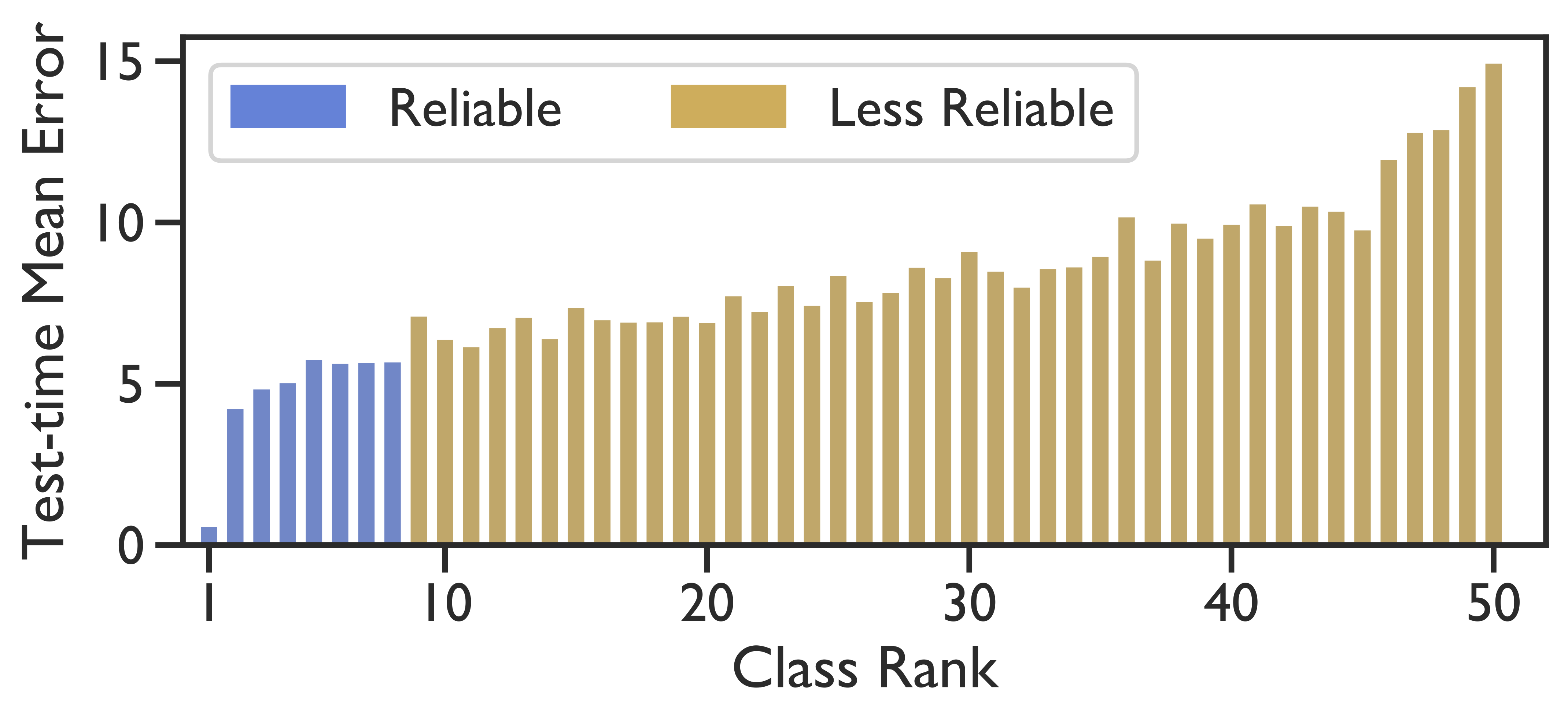}
    \vspace{-3mm}
    \caption{The test-time mean error increases with the rank of the class. Classes with lower ranks are more robust to test-time distribution shift (lower errors).}
    \label{fig:rank}
    \vspace{-3mm}
\end{figure}

\subsection{Principal Entropy Minimization}
Entropy minimization is a commonly used technique in test-time adaptation \cite{wang2020tent, liu2021source, lee2023towards}, as it does not require the labels to compute loss objective. Although it improves performance, entropy minimization inevitably introduce noisy gradient signals, and there are some works that tackle this problem from the sample perspective \cite{zhang2023domainadaptor, gao2024unified, xiong2024robust}. However, these methods may fail to fully utilize all the samples from the test data. In this part, we introduce principal entropy minimization that tackles this problem from the class perspective. Specifically, we can write the entropy of $\bm p$ as:
\begin{equation}
\label{eq:original-entropy}
\mathcal H(\bm p) = -\sum_{i\in \mathcal S} p_i \log  p_i,
\end{equation}
where $\mathcal S=\{1, 2, \cdots, C\}$ is the set of all classes. 

However, Eq. \ref{eq:original-entropy} contains terms of all classes, including the more reliable ones and the less reliable ones. Denote the rank of each class as $r_i, i\in \mathcal S$, which can be formally defined as:
\begin{equation}
r_i=|\{p_j ~|~ j\in \mathcal S \land p_j\ge p_i\}|,
\end{equation}
where $|\cdot|$ denotes the cardinality of a set. Our observation is that classes with lower ranks are more reliable. The empirical evidence is presented in Figure \ref{fig:rank}, where we measure the changes of predicted probability with respect to the rank of the class. The results show that classes with lower ranks (or relatively higher probabilities) are more robust to test-time distribution shifts. Therefore, it is reasonable to exclude the less reliable set from $\mathcal S$ in the computation of entropy.

Specifically, we define the reliable class set $\mathcal S_R^{(k)}$ \textit{for each test sample} based on the ranks $r_i$ derived from the predicted probabilities $p_i$, which is defined as follows:
\begin{equation}\label{eq:rank-set}
\mathcal S_R^{(k)} = \{i \in \mathcal S |r_i\le k\},
\end{equation}
where $k$ is a hyper-parameter. Subsequently, we define the principal entropy of $\bm p$ as:
\begin{equation} \label{eq:principal-entropy}
\mathcal H_P^{(k)}(\bm p) = -\sum_{i\in \mathcal S_R^{(k)}} p_i \log  p_i.
\end{equation}
The principal entropy is then used as the minimizing objective to replace the entropy defined in Eq. \ref{eq:original-entropy}:
\begin{equation}\label{eq:loss-pem}
\mathcal L_{PEM} = \mathcal H_P^{(k)}(\bm p).
\end{equation}

\begin{algorithm}[tb]
    \caption{Optimization Algorithm of \method{}}
    \label{alg:algorithm}
    \textbf{Requires}: The well-trained model $\mathcal M=(\mathcal E^A, \mathcal E^V, \mathcal F)$, the unlabeled test dataset $\mathcal D_{te}$, and $k$ in Eq. \ref{eq:rank-set}.\\
    \textbf{Ensures}: The adapted model, and the prediction on test data.

    \begin{algorithmic}[1] 
    \FOR{each batch in $\mathcal D_{te}$}
    \STATE Compute the embeddings of inputs, \emph{i.e.} $\{ \bm z_i^A \}_{i=1}^{T_A} = \mathcal E^A(\bm x^A)$ and $\{\bm z_i^V \}_{i=1}^{T_V}= \mathcal E^V(\bm x^V)$\\
    \STATE Compute the attention scores using Eq. \ref{eq:unnormalized-attention}.\\
    \STATE Compute the attention bootstrapping loss using Eq. \ref{eq:loss-ab}.\\
    \STATE Obtain the predicted probabilities $\bm p$ from the output of the fusion module $\mathcal F$.\\
    \STATE Sort the predicted probabilities to obtain the ranks $r_i$ defined in Eq. \ref{eq:rank-set}.\\
    \STATE Compute the principal entropy in Eq. \ref{eq:principal-entropy} as $\mathcal L_{PEM}$.\\
    \STATE Compute the final loss function in Eq. \ref{eq:final-loss}.
    \STATE Update the tunable parameters in the fusion module $\mathcal F$ through back-propagation.
    \ENDFOR
    \end{algorithmic}
\end{algorithm}

\subsection{Summary}
In this part, we provide a summary of our method. When the multi-modal learning system $\mathcal M$ receives data, it first encodes the inputs of each modality using modality-specific encoders $\mathcal E^A$ and $\mathcal E^V$. Then the learned embeddings are sent into the fusion module $\mathcal F$, in which attention bootstrapping is performed using the attention scores. The fusion module yields a probability distribution for each sample, and the principal entropy is computed as the loss objective. 
The final loss function can be written as:
\begin{equation}
\label{eq:final-loss}
\mathcal L = \lambda\mathcal L_{AB} + \mathcal L_{PEM}
\end{equation}
The optimization procedure is summarized in Algorithm \ref{alg:algorithm}. It can be shown that this algorithm has the same time complexity as the model $\mathcal M$ without adaptation, which is $\mathcal O(N_{te}d (T_A+T_V)^2)$. 
Empirical results about efficiency are provided in the experiment section.
\section{Experiments}
\subsection{Experimental Settings}
\label{sec:exp-settings}
\noindent\textbf{Benchmarks.} 
The experiments are performed on two benchmarks: Kinetics50-C and VGGSound-C \cite{yang2024test}, which are based on the widely used Kinetics \cite{kay2017kinetics} and VGGSound \cite{chen2020vggsound} datasets. Each of the benchmarks contains two settings, \emph{i.e.} corrupted video setting (which contains 15 types of video corruptions) and corrupted audio setting (which contains 6 types of audio corruptions). Each type of corruption has 5 severity, and we adopt severity 5 as default following \cite{yang2024test}. In the Kinetics50 dataset (from which Kinetics50-C benchmark is constructed), the video modality typically contains more information, whereas in the VGGSound dataset (from which VGGSound-C is constructed), the audio modality typically contains more information.

\smallskip
\noindent\textbf{Baselines Methods.} The proposed \method{} is compared with several competing baselines, including Tent \cite{wang2020tent}, MMT \cite{shin2022mm}, EATA \cite{niu2022efficient}, SAR \cite{niu2023towards}, and READ \cite{yang2024test}. 

\smallskip
\noindent\textbf{Implementation Details.} In the experiments, we use CAV-MAE \cite{gong2023contrastive} as the architecture of $\mathcal M$. The model is pretrained on the corresponding training set (Kinetics or VGGSound). We set $k$ in Eq. \ref{eq:rank-set} to about 8 for Kinetics50-C and 30 for  VGGSound-C, and $\lambda$ to 1 by default. Moreover, we also use a class-balancing loss in alignment with \cite{yang2024test}. For optimization, we use Adam optimizer \cite{kingma2014adam} and the model is optimized within a single epoch, with the learning rate of $1\times 10^{-4}$.

\begin{table*}[ht]
\centering
\resizebox{\textwidth}{!}{%
\begin{tabular}{c ccc c cccc c cccc c cccc c}
\toprule
\multirow{2}{*}{Models} & \multicolumn{3}{c}{Noise} && \multicolumn{4}{c}{Blur} && \multicolumn{4}{c}{Weather} && \multicolumn{4}{c}{Digital} & \multirow{2}{*}{Avg.} \\
\cmidrule{2-4} \cmidrule{6-9} \cmidrule{11-14} \cmidrule{16-19} & Gauss. & Shot & Impul. && Defoc. & Glass & Mot. & Zoom && Snow & Frost & Fog & Brit. && Contr. & Elas. & Pix. & JPEG &\\
\midrule
Raw & 46.8 & 48.0 & 46.9 && 67.5 & 62.2 & 70.6 & 67.7 && 61.6 & 60.3 & 46.7 & 75.2 && 52.1 & 65.7 & 66.5 & 61.9 & 59.9 \\
MMT & 46.2 & 46.6 & 46.1 && 58.8 & 55.7 & 62.4 & 61.7 && 52.6 & 54.4 & 48.5 & 69.3 && 49.3 & 57.6 & 56.4 & 54.5 & 54.5 \\
Tent & 46.3 & 47.0 & 46.3 && 67.4 & 62.5 & 70.4 & 67.7 && 63.1 & 61.1 & 34.9 & 75.4 && 51.6 & 66.7 & 66.5 & 62.0 & 59.4 \\
EATA & 46.8 & 47.6 & 47.1 && 67.2 & 61.8 & 70.2 & 67.7 && 61.6 & 60.6 & 46.0 & 75.2 && 52.4 & 65.9 & 66.4 & 62.7 & 60.1 \\
SAR & 46.7 & 47.4 & 46.6 && 67.0 & 61.7 & 70.0 & 66.4 && 61.8 & 60.6 & 46.0 & 75.2 && 52.1 & 65.7 & 66.0 & 62.0 & 59.8 \\
READ & 49.4 & 49.7 & 49.0 && 68.0 & 65.1 & 71.2 & 69.0 && 64.5 & 64.4 & 57.4 & 75.5 && 53.6 & 68.3 & 68.0 & 65.1 & 62.5 \\
\midrule
\textbf{\method{}} & \textbf{50.3} & \textbf{51.1} & \textbf{50.4} && \textbf{70.0} & \textbf{69.6} & \textbf{72.5} & \textbf{71.2} && \textbf{65.2} & \textbf{66.2} & \textbf{65.6} & \textbf{75.7} && \textbf{56.6} & \textbf{71.9} & \textbf{70.5} & \textbf{67.8} & \textbf{65.0} \\
\bottomrule
\end{tabular}}
\vspace{-1mm}
\caption{Prediction accuracies (in \%) on Kinetics50-C benchmark (corrupted video modality).}
\label{tab:ks50v}
\end{table*}
\begin{table*}[ht]
\centering
\resizebox{0.95\textwidth}{!}{%
\begin{tabular}{c ccc c ccc c ccc c ccc c}
\toprule
\multirow{2}{*}{Models} & \multicolumn{3}{c}{Noise} && \multicolumn{3}{c}{Weather} & \multirow{2}{*}{Avg.} & \multicolumn{3}{c}{Noise} && \multicolumn{3}{c}{Weather} & \multirow{2}{*}{Avg.} \\
\cmidrule{2-4} \cmidrule{6-8} \cmidrule{10-12} \cmidrule{14-16}
& Gauss. & Traff. & Crowd. && Rain & Thund. & Wind & & Gauss. & Traff. & Crowd. && Rain & Thund. & Wind & \\
\midrule
Raw & 73.7 & 65.5 & 67.9 && 70.3 & 67.9 & 70.3 & 69.3 & 37.0 & 25.5 & 16.8 && 21.6 & 27.3 & 25.5 & 25.6 \\
MMT & 70.8 & 69.2 & 68.5 && 69.0 & 69.8 & 68.5 & 69.4 & 14.1 & 5.2 & 6.4 && 9.8 & 8.6 & 4.5 & 7.6 \\
Tent & 73.9 & 67.4 & 68.5 && 70.4 & 66.5 & 70.4 & 69.6 & 10.6 & 2.6 & 1.8 && 2.3 & 3.3 & 4.1 & 4.5 \\
EATA & 73.7 & 66.1 & 68.5 && 69.5 & 70.6 & 69.4 & 69.4 & 39.2 & 26.1 & 22.9 && 26.0 & 31.7 & 30.4 & 29.4 \\ 
SAR & 73.7 & 65.4 & 68.2 && 69.9 & 67.2 & 70.2 & 69.1 & 37.4 & 9.5 & 11.0 && 12.1 & 26.8 & 23.7 & 20.1 \\ 
READ & 74.1 & 69.0 & 69.7 && 71.1 & 71.8 & 70.7 & 71.1 & 40.4 & 28.9 & 26.6 && 30.9 & 36.7 & 30.6 & 32.4 \\
\midrule
\textbf{\method{}} & \textbf{74.8} & \textbf{71.3} & \textbf{71.5} && \textbf{71.9} & \textbf{73.8} & \textbf{71.6} & \textbf{72.5} & \textbf{40.6} & \textbf{33.7} & \textbf{34.8} && \textbf{32.2} & \textbf{41.1} & \textbf{34.4} & \textbf{36.1} \\
\bottomrule
\end{tabular}}
\vspace{-1mm}
\caption{Prediction accuracies (in \%) on Kinetics50-C (left) and VGGSound-
C (right) benchmarks (corrupted audio modality).}
\label{tab:audio}
\end{table*}
\begin{table*}[h!]
\centering
\resizebox{\textwidth}{!}{%
\begin{tabular}{c ccc c cccc c cccc c cccc c}
\toprule
\multirow{2}{*}{Models} & \multicolumn{3}{c}{Noise} && \multicolumn{4}{c}{Blur} && \multicolumn{4}{c}{Weather} && \multicolumn{4}{c}{Digital} & \multirow{2}{*}{Avg.} \\
\cmidrule{2-4} \cmidrule{6-9} \cmidrule{11-14} \cmidrule{16-19} & Gauss. & Shot & Impul. && Defoc. & Glass & Mot. & Zoom && Snow & Frost & Fog & Brit. && Contr. & Elas. & Pix. & JPEG &\\
\midrule
Raw & 52.8 & 52.7 & 52.7 && 57.2 & 57.2 & 58.7 & 56.8 && 56.4 & 56.6 & 55.6 & 58.9 && 53.7 & 56.9 & 55.8 & 56.9 & 56.0 \\
MMT & 7.1 & 7.3 & 7.3 && 44.8 & 41.5 & 48.0 & 45.5 && 27.4 & 23.5 & 30.5 & 46.3 && 24.0 & 43.0 & 40.7 & 45.7 & 32.0 \\
Tent & 52.7 & 52.7 & 52.7 && 56.7 & 56.5 & 58.0 & 56.5 && 55.0 & 57.0 & 56.3 & 58.7 && 54.0 & 57.4 & 56.7 & 57.4 & 55.8 \\
EATA & 53.0 & 52.8 & 53.0 && 57.2 & 57.1 & 58.6 & 57.8 && 56.3 & 56.8 & 56.4 & 59.0 && 54.1 & 57.4 & 56.1 & 57.0 & 56.2 \\
SAR & 52.9 & 52.8 & 52.9 && 57.0 & 57.1 & 58.5 & 56.8 && 56.3 & 56.7 & 55.9 & 58.9 && 54.0 & 57.6 & 57.1 & 57.2 & 56.1 \\
READ & 53.6 & 53.6 & 53.5 && 57.9 & 57.7 & 59.4 & 58.8 && 57.2 & 57.8 & 55.0 & 59.9 && 55.2 & 58.6 & 57.1 & 57.9 & 56.9 \\
\midrule
\textbf{\method{}} & \textbf{54.0} & \textbf{53.9} & \textbf{54.0} && \textbf{58.2} & \textbf{58.1} & \textbf{59.6} & \textbf{59.3} && \textbf{57.5} & \textbf{58.2} & \textbf{58.2} & \textbf{60.2} && \textbf{56.2} & \textbf{59.1} & \textbf{57.5} & \textbf{58.3} & \textbf{57.5} \\
\bottomrule
\end{tabular}}
\vspace{-1mm}
\caption{Prediction accuracies (in \%) on VGGSound-C benchmark (corrupted video modality).}
\vspace{-3mm}
\label{tab:vggv}
\end{table*}

\subsection{Performance Comparison}
We first compare the performance of the proposed \method{} and the baselines in Table \ref{tab:ks50v}, Table \ref{tab:audio} and Table \ref{tab:vggv}. 
The first model (Raw) denotes the model without any test-time adaptation. 
From the results, we have several observations.

\begin{itemize}
    \item The proposed \method{} achieves a consistent lead in both Kinetics50-C and VGGSound-C benchmarks in the face of various types of test-time distribution shifts. This shows the overall effectiveness of the proposed \method{}.
    \item Our model experiences more significant improvement when facing the test-time distribution that affects the more informative modality (\emph{i.e.} corrupted video modality of Kinetics50-C and corrupted audio modality of VGGSound). Previous efforts often ignore the problem of increasing attention gap, whereas the proposed \method{} explicitly reduces this gap, which is beneficial for the model to incorporate the more informative modality with the other modality, and thus achieves higher accuracy.
    \item The task of multi-modal test-time adaptation is inherently hard. When the model is challenged by adverse distribution shifts, some models fail to achieve satisfactory performance, and sometimes even worse than the model before adaptation. This shows that without ground truth labels, the gradients can be very noisy and are potentially harmful for the model. Our model adopts principal entropy minimization, which reduces the noise in the gradient and leads to better results.
\end{itemize}

\begin{figure}
    \centering
    \includegraphics[width=\linewidth]{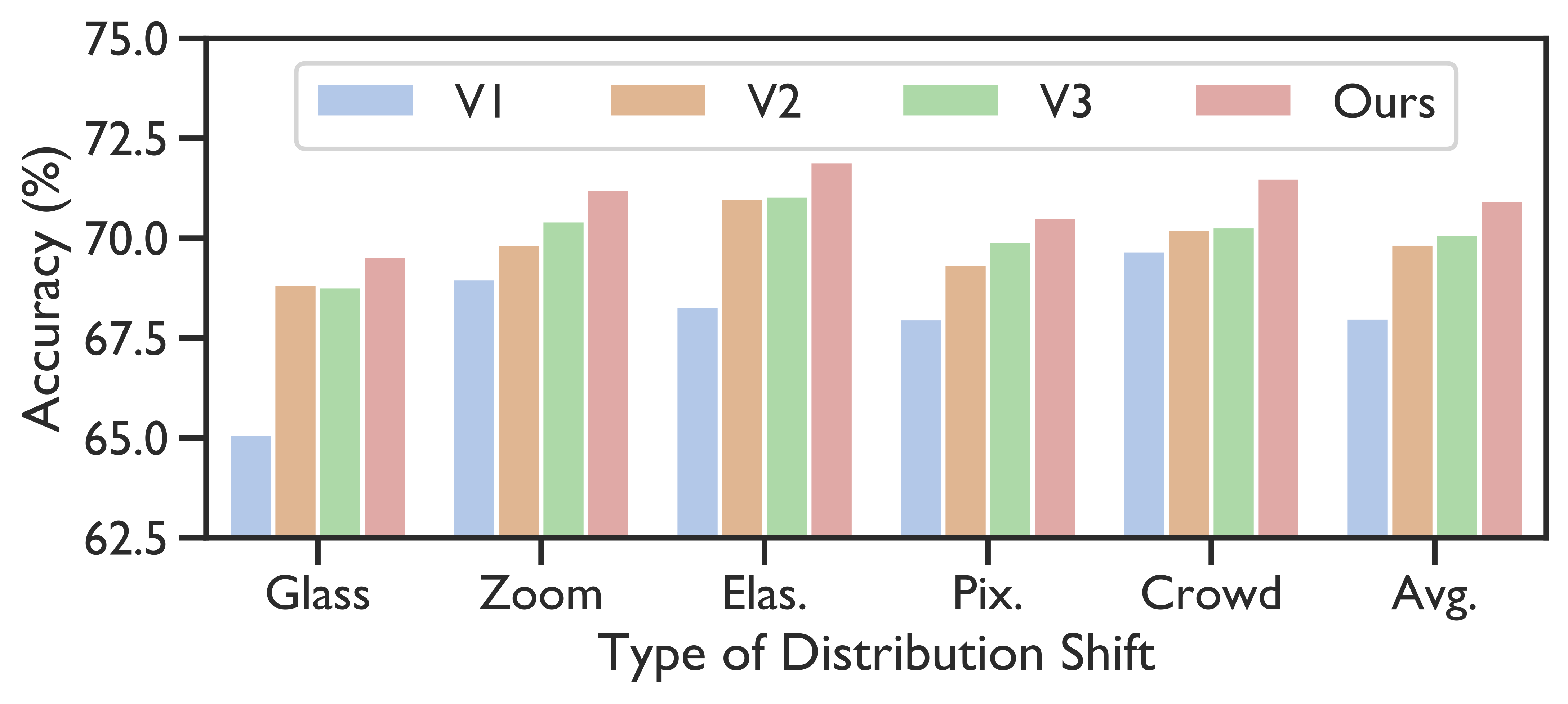}
    \vspace{-6mm}
    \caption{Ablation of the main components of \method{}.}
    \vspace{-2mm}
    \label{fig:abl}
\end{figure}
\subsection{Ablation Studies}
\textbf{Ablation of the main components.}
We design several variants of the model to investigate the role of attention bootstrapping and principal entropy minimization. V1 is the model that adopt a tunable layer 
, and use a basic self-supervised objective in \cite{yang2024test}. V2 is the model that uses attention bootstrapping. V3 is the model that replaces principal entropy minimization with vanilla entropy minimization. The last model is the proposed \method{}, which contains both attention bootstrapping and principal entropy minimization. The results on Kinetics50-C benchmark are shown in Figure \ref{fig:abl}. As can be seen from the results, the use of attention bootstrapping increases accuracy significantly (comparing V1 and V2), this can be attributed to the better fusion of modalities under distribution shift. Moreover, the improvement of vanilla entropy minimization is marginal (comparing V2 and V3), whereas the proposed principal entropy minimization further boosts the accuracy (comparing V2 and Ours). This shows that reducing the noise in the gradients, which is achieved by principal entropy minimization, is beneficial for the performance.

\smallskip
\noindent\textbf{Ablation of $k$ in Eq. \ref{eq:rank-set}.} We then investigate the role of hyperparameter $k$ in principal entropy minimization. The results on Kinetics50-C benchmark are shown in Table \ref{tab:sensitivity}. As can be seen from the table, the model is generally not sensitive to $k$, and the highest accuracy is achieved at 8. When $k$ is small, the reliable class set $\mathcal S_R^{(k)}$ for each sample is small, which may not fully utilize all the information. Conversely, when $k$ is large, the reliability of class probabilities $p_i$ decreases, leading to noisy gradient information.

\begin{table}[ht]
    \centering
    \resizebox{\linewidth}{!}{%
    \begin{tabular}{c cccccc}
    \toprule
    Models & Raw & Tent & EATA & SAR & READ & \method{} \\
    \midrule
    Samples per second & 92.4 & 68.5 & 69.8 & 55.6 & 88.2 & 87.3 \\
    \midrule
    \# Tunable parameters & 0 & 0.2M & 0.2M & 0.2M & 1.8M & 1.8M
    \\
    \bottomrule
    \end{tabular}}
    \caption{Comparison of models' efficiency.}
    \label{tab:eff}
\end{table}
\begin{table}[t]
    \centering
    \resizebox{\linewidth}{!}{%
    \begin{tabular}{c cccccc}
    \toprule
    $k$ & Glass & Zoom & Elas. & Pix. & Crowd & Avg. \\
    \midrule
    6 & 69.28 & 70.83 & 71.71 & 69.88 & 71.49 & 70.64 \\
    7 & 69.40 & 71.18 & 71.89 & 70.42 & 71.48 & 70.87 \\
    8 & 69.56 & 71.24 & 71.93 & 70.53 & 71.52 & 70.96 \\
    9 & 69.53 & 71.33 & 71.83 & 70.59 & 71.49 & 70.95\\
    10 & 69.55 & 71.34 & 71.75 & 70.61 & 71.48 & 70.94 \\
    \bottomrule
    \end{tabular}}
    \caption{Ablated study about $k$ in Eq. \ref{eq:rank-set}.}
    \label{tab:sensitivity}
\end{table}
\begin{figure}[!h]
    \centering
    \includegraphics[width=\linewidth]{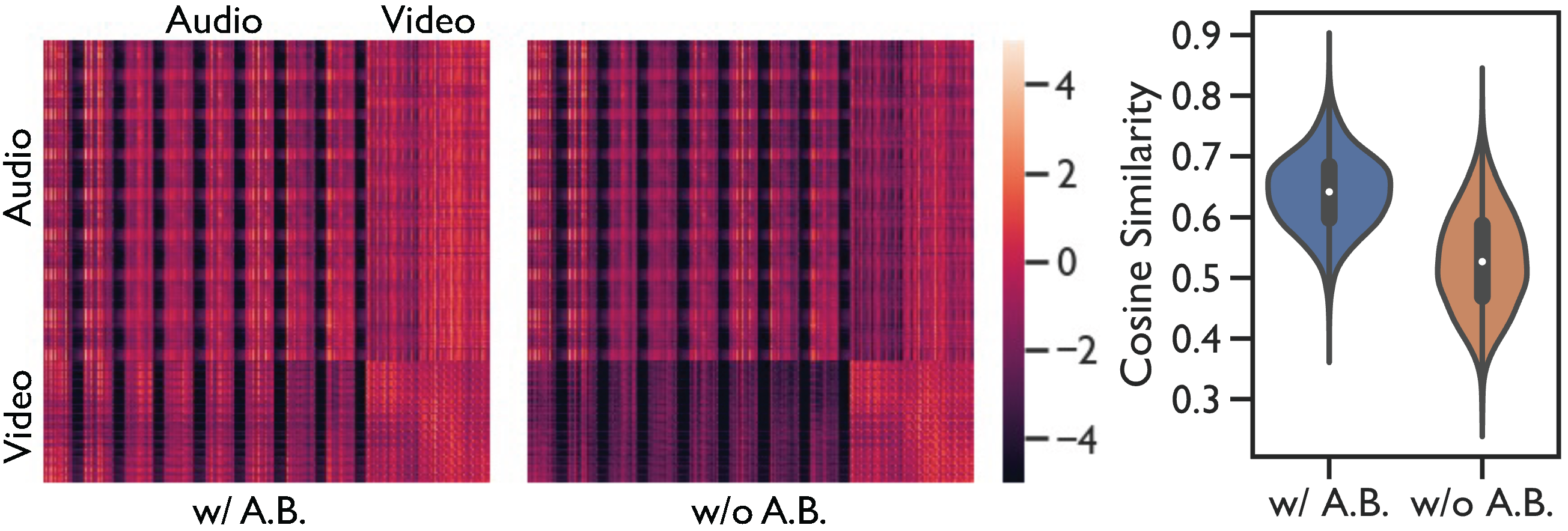}
    \vspace{-2mm}
    \caption{The attention map with attention bootstrapping (w/ A.B., left) and without (w/o A.B., middle). The proposed attention bootstrapping encourages cross-attention, and achieves better alignment with increased cosine similarity between token embeddings of different modalities.}
    \vspace{-2mm}
    \label{fig:abmap}
\end{figure}

\subsection{Efficiency Comparison}
We then show that the proposed \method{} is efficient. As mentioned in the previous section, the time complexity of the model is equivalent to the raw model $\mathcal M$ without adaptation. In Table \ref{tab:eff}, empirical evidences are provided where we count the number of samples that the model processes per second and the total number of tunable parameters on the Kinetics50-C dataset. As is shown in the table, \method{} achieves similar speed (samples per second) compared to READ. Moreover, it is faster than methods that require tuning layer normalization modules (Tent, EATA, and SAR) with relatively more tunable parameters, similar to READ. The results show the efficiency of the proposed method.

\subsection{Further Analysis}
\textbf{Better alignment with decreased attention gap.} We investigate the role of attention bootstrapping in the alignment and fusion of different modalities. As mentioned before, attention bootstrapping encourages cross-attention with the help of self-attention, and we visualize the unnormalized attention map $\tilde{\bm A}$ of Eq. \ref{eq:unnormalized-attention} in Figure \ref{fig:abmap} (left and middle, on the Kinetics50-C benchmark, Fog corruption). The results show a significant increase in cross-attention scores, and the attention gap $\mu_{V2V}-\mu_{A2V}$ reduces from 1.02 to 0.11 (another gap $\mu_{A2A}-\mu_{V2A}$ reduces from 0.98 to 0.23). This leads to more aligned representations, as shown in Figure \ref{fig:abmap} (right), where we plot the distribution of cosine similarities between the token embeddings of two modalities. This shows that the representations of different modalities are more aligned (higher similarity) when attention bootstrapping is used. 

\begin{figure}
    \centering
    \includegraphics[width=\linewidth]{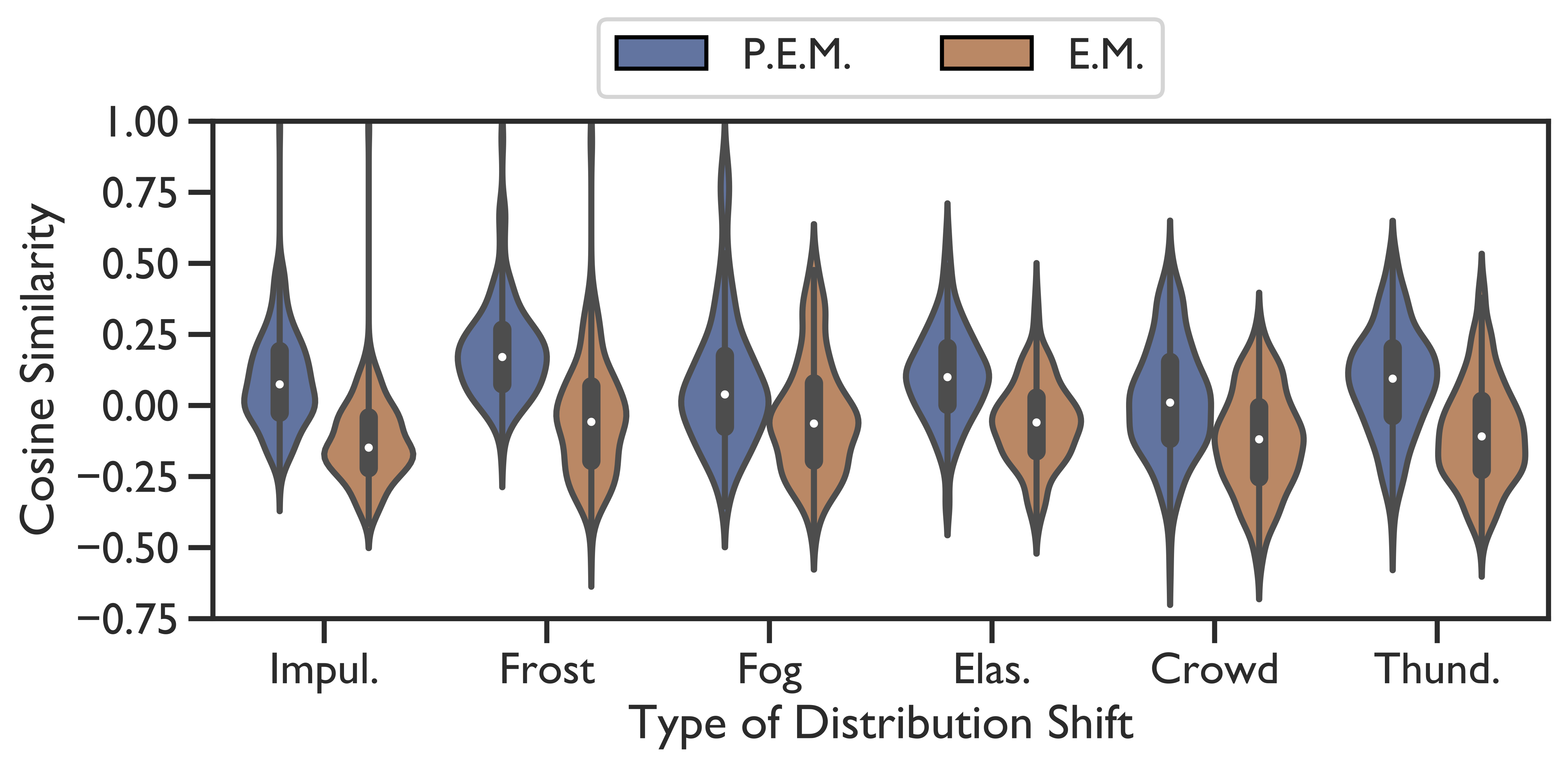}
    \vspace{-2mm}
    \caption{The cosine similarities between the gradients of cross entropy objective (with ground truth labels) and the gradients of principal entropy minimization (P.E.M.) / entropy minimization (E.M.) objective. P.E.M. yields gradient directions closer to the ground truth than E.M.}
    \vspace{-2mm}
    \label{fig:pem-violin}
\end{figure}

\smallskip
\noindent\textbf{Reduced gradient noise.}
We then show the effect of principal entropy minimization in the reduction of gradient noise. Specifically, we perform experiments on Kinetics50-C under various types of distribution shifts and compare the gradients generated from two sources: (1) the cross entropy loss objective using the ground truth labels of the samples, and (2) the principal entropy minimization (P.E.M.) objective or the commonly used entropy minimization (E.M.) objective. The distributions of the cosine similarities between gradients from sources (1) and (2) are illustrated in Figure \ref{fig:pem-violin}. As can be seen from the figure, the gradients generated by P.E.M. objective are closer to the gradients generated using ground truth labels. This shows that by excluding the less reliable classes from the computation of entropy, P.E.M. objective reduces the noise signals in the gradients.

\section{Conclusion}
This paper tackles the problem of multi-modal test-time adaptation, and proposes attention bootstrapping and principal entropy minimization (\method{}) to solve this problem. When the multi-modal learning system is influenced by distribution shifts, modality mismatch occurs, and the attention gap increases, which hinders the fusion of different modalities. To reduce this gap, attention bootstrapping is proposed. Moreover, we observe that classes with lower probabilities are less reliable and may introduce noise in the gradients. To tackle this, we differentiate the reliable classes and less reliable classes with the proposed principal entropy minimization. Extensive experiments on the benchmark datasets demonstrate the effectiveness of the proposed \method{}.

\section*{Acknowledgments}
This paper is partially supported by the National Key Research and Development Program of China with Grant No. 2023YFC3341203 as well as the National Natural Science Foundation of China with Grant Number 62276002.

\bibliography{0_main}

\begin{thebibliography}{68}
\providecommand{\natexlab}[1]{#1}

\bibitem[{Azimi et~al.(2022)Azimi, Palacio, Raue, Hees, Bertinetto, and Dengel}]{azimi2022self}
Azimi, F.; Palacio, S.; Raue, F.; Hees, J.; Bertinetto, L.; and Dengel, A. 2022.
\newblock Self-supervised test-time adaptation on video data.
\newblock In \emph{WACV}, 3439--3448.

\bibitem[{Blikstein(2013)}]{blikstein2013multimodal}
Blikstein, P. 2013.
\newblock Multimodal learning analytics.
\newblock In \emph{LAK}, 102--106.

\bibitem[{Boudiaf et~al.(2022)Boudiaf, Mueller, Ben~Ayed, and Bertinetto}]{boudiaf2022parameter}
Boudiaf, M.; Mueller, R.; Ben~Ayed, I.; and Bertinetto, L. 2022.
\newblock Parameter-free online test-time adaptation.
\newblock In \emph{CVPR}, 8344--8353.

\bibitem[{Chen et~al.(2022)Chen, Wang, Darrell, and Ebrahimi}]{chen2022contrastive}
Chen, D.; Wang, D.; Darrell, T.; and Ebrahimi, S. 2022.
\newblock Contrastive test-time adaptation.
\newblock In \emph{CVPR}, 295--305.

\bibitem[{Chen et~al.(2020)Chen, Xie, Vedaldi, and Zisserman}]{chen2020vggsound}
Chen, H.; Xie, W.; Vedaldi, A.; and Zisserman, A. 2020.
\newblock Vggsound: A large-scale audio-visual dataset.
\newblock In \emph{ICASSP}, 721--725. IEEE.

\bibitem[{Chen et~al.(2023)Chen, Zhang, Song, Shan, and Liu}]{chen2023improved}
Chen, L.; Zhang, Y.; Song, Y.; Shan, Y.; and Liu, L. 2023.
\newblock Improved test-time adaptation for domain generalization.
\newblock In \emph{CVPR}, 24172--24182.

\bibitem[{Fan et~al.(2023)Fan, Xu, Wang, Wang, and Guo}]{fan2023pmr}
Fan, Y.; Xu, W.; Wang, H.; Wang, J.; and Guo, S. 2023.
\newblock Pmr: Prototypical modal rebalance for multimodal learning.
\newblock In \emph{CVPR}, 20029--20038.

\bibitem[{Gao et~al.(2023)Gao, Zhang, Liu, Darrell, Shelhamer, and Wang}]{gao2023back}
Gao, J.; Zhang, J.; Liu, X.; Darrell, T.; Shelhamer, E.; and Wang, D. 2023.
\newblock Back to the source: Diffusion-driven adaptation to test-time corruption.
\newblock In \emph{CVPR}, 11786--11796.

\bibitem[{Gao, Zhang, and Liu(2024)}]{gao2024unified}
Gao, Z.; Zhang, X.-Y.; and Liu, C.-L. 2024.
\newblock Unified Entropy Optimization for Open-Set Test-Time Adaptation.
\newblock In \emph{CVPR}, 23975--23984.

\bibitem[{Gong et~al.(2023)Gong, Rouditchenko, Liu, Harwath, Karlinsky, Kuehne, and Glass}]{gong2023contrastive}
Gong, Y.; Rouditchenko, A.; Liu, A.~H.; Harwath, D.; Karlinsky, L.; Kuehne, H.; and Glass, J.~R. 2023.
\newblock Contrastive Audio-Visual Masked Autoencoder.
\newblock In \emph{ICLR}.

\bibitem[{He et~al.(2021)He, Zhao, Luo, Hui, Huang, Zhang, and Liu}]{he2021transrefer3d}
He, D.; Zhao, Y.; Luo, J.; Hui, T.; Huang, S.; Zhang, A.; and Liu, S. 2021.
\newblock Transrefer3d: Entity-and-relation aware transformer for fine-grained 3d visual grounding.
\newblock In \emph{ACM MM}, 2344--2352.

\bibitem[{Hu et~al.(2021)Hu, Uzunbas, Chen, Wang, Shah, Nevatia, and Lim}]{hu2021mixnorm}
Hu, X.; Uzunbas, G.; Chen, S.; Wang, R.; Shah, A.; Nevatia, R.; and Lim, S.-N. 2021.
\newblock Mixnorm: Test-time adaptation through online normalization estimation.
\newblock \emph{arXiv preprint arXiv:2110.11478}.

\bibitem[{Huang et~al.(2024)Huang, Chen, Guo, Zeng, Zhao, Wu, Yuan, Zhao, Guo, Zhang et~al.}]{huang2024mmevalpro}
Huang, J.; Chen, L.; Guo, T.; Zeng, F.; Zhao, Y.; Wu, B.; Yuan, Y.; Zhao, H.; Guo, Z.; Zhang, Y.; et~al. 2024.
\newblock Mmevalpro: Calibrating multimodal benchmarks towards trustworthy and efficient evaluation.
\newblock \emph{arXiv preprint arXiv:2407.00468}.

\bibitem[{Karmanov et~al.(2024)Karmanov, Guan, Lu, El~Saddik, and Xing}]{karmanov2024efficient}
Karmanov, A.; Guan, D.; Lu, S.; El~Saddik, A.; and Xing, E. 2024.
\newblock Efficient Test-Time Adaptation of Vision-Language Models.
\newblock In \emph{CVPR}, 14162--14171.

\bibitem[{Kay et~al.(2017)Kay, Carreira, Simonyan, Zhang, Hillier, Vijayanarasimhan, Viola, Green, Back, Natsev et~al.}]{kay2017kinetics}
Kay, W.; Carreira, J.; Simonyan, K.; Zhang, B.; Hillier, C.; Vijayanarasimhan, S.; Viola, F.; Green, T.; Back, T.; Natsev, P.; et~al. 2017.
\newblock The kinetics human action video dataset.
\newblock \emph{arXiv preprint arXiv:1705.06950}.

\bibitem[{Kingma and Ba(2014)}]{kingma2014adam}
Kingma, D.~P.; and Ba, J. 2014.
\newblock Adam: A method for stochastic optimization.
\newblock \emph{arXiv preprint arXiv:1412.6980}.

\bibitem[{Krauhausen et~al.(2024)Krauhausen, Griggs, McCulloch, den Toonder, Gkoupidenis, and van~de Burgt}]{krauhausen2024bio}
Krauhausen, I.; Griggs, S.; McCulloch, I.; den Toonder, J.~M.; Gkoupidenis, P.; and van~de Burgt, Y. 2024.
\newblock Bio-inspired multimodal learning with organic neuromorphic electronics for behavioral conditioning in robotics.
\newblock \emph{Nature Communications}, 15(1): 4765.

\bibitem[{Kundu et~al.(2020)Kundu, Venkat, Babu et~al.}]{kundu2020universal}
Kundu, J.~N.; Venkat, N.; Babu, R.~V.; et~al. 2020.
\newblock Universal source-free domain adaptation.
\newblock In \emph{CVPR}, 4544--4553.

\bibitem[{Lee et~al.(2023{\natexlab{a}})Lee, Ahuja, Liang, Natu, and Morency}]{lee2023lecture}
Lee, D.~W.; Ahuja, C.; Liang, P.~P.; Natu, S.; and Morency, L.-P. 2023{\natexlab{a}}.
\newblock Lecture presentations multimodal dataset: Towards understanding multimodality in educational videos.
\newblock In \emph{ICCV}, 20087--20098.

\bibitem[{Lee et~al.(2023{\natexlab{b}})Lee, Das, Choo, and Choi}]{lee2023towards}
Lee, J.; Das, D.; Choo, J.; and Choi, S. 2023{\natexlab{b}}.
\newblock Towards open-set test-time adaptation utilizing the wisdom of crowds in entropy minimization.
\newblock In \emph{ICCV}, 16380--16389.

\bibitem[{Lee et~al.(2024)Lee, Jung, Lee, Park, Shin, Hwang, and Yoon}]{lee2024entropy}
Lee, J.; Jung, D.; Lee, S.; Park, J.; Shin, J.; Hwang, U.; and Yoon, S. 2024.
\newblock Entropy is not enough for test-time adaptation: From the perspective of disentangled factors.
\newblock \emph{arXiv preprint arXiv:2403.07366}.

\bibitem[{Liang, He, and Tan(2024)}]{liang2024comprehensive}
Liang, J.; He, R.; and Tan, T. 2024.
\newblock A comprehensive survey on test-time adaptation under distribution shifts.
\newblock \emph{IJCV}, 1--34.

\bibitem[{Lim et~al.(2023)Lim, Kim, Choo, and Choi}]{lim2023ttn}
Lim, H.; Kim, B.; Choo, J.; and Choi, S. 2023.
\newblock TTN: A domain-shift aware batch normalization in test-time adaptation.
\newblock \emph{arXiv preprint arXiv:2302.05155}.

\bibitem[{Litrico, Del~Bue, and Morerio(2023)}]{litrico2023guiding}
Litrico, M.; Del~Bue, A.; and Morerio, P. 2023.
\newblock Guiding pseudo-labels with uncertainty estimation for source-free unsupervised domain adaptation.
\newblock In \emph{CVPR}, 7640--7650.

\bibitem[{Liu et~al.(2021)Liu, Kothari, Van~Delft, Bellot-Gurlet, Mordan, and Alahi}]{liu2021ttt++}
Liu, Y.; Kothari, P.; Van~Delft, B.; Bellot-Gurlet, B.; Mordan, T.; and Alahi, A. 2021.
\newblock Ttt++: When does self-supervised test-time training fail or thrive?
\newblock \emph{NeurIPS}, 34: 21808--21820.

\bibitem[{Liu et~al.(2023)Liu, Qiao, Lu, Yin, Lin, Peng, and Ren}]{liu2023osan}
Liu, Y.; Qiao, L.; Lu, C.; Yin, D.; Lin, C.; Peng, H.; and Ren, B. 2023.
\newblock OSAN: A one-stage alignment network to unify multimodal alignment and unsupervised domain adaptation.
\newblock In \emph{CVPR}, 3551--3560.

\bibitem[{Liu, Zhang, and Wang(2021)}]{liu2021source}
Liu, Y.; Zhang, W.; and Wang, J. 2021.
\newblock Source-free domain adaptation for semantic segmentation.
\newblock In \emph{CVPR}, 1215--1224.

\bibitem[{Ma et~al.(2024{\natexlab{a}})Ma, Zhu, Zhang, Zhao, Wu, Huang, Hu, and Wu}]{ma2024invariant}
Ma, H.; Zhu, Y.; Zhang, C.; Zhao, P.; Wu, B.; Huang, L.-K.; Hu, Q.; and Wu, B. 2024{\natexlab{a}}.
\newblock Invariant Test-Time Adaptation for Vision-Language Model Generalization.
\newblock \emph{arXiv preprint arXiv:2403.00376}.

\bibitem[{Ma(2024)}]{ma2024improved}
Ma, J. 2024.
\newblock Improved Self-Training for Test-Time Adaptation.
\newblock In \emph{CVPR}, 23701--23710.

\bibitem[{Ma et~al.(2022)Ma, Ren, Zhao, Testuggine, and Peng}]{ma2022multimodal}
Ma, M.; Ren, J.; Zhao, L.; Testuggine, D.; and Peng, X. 2022.
\newblock Are multimodal transformers robust to missing modality?
\newblock In \emph{CVPR}, 18177--18186.

\bibitem[{Ma et~al.(2021)Ma, Ren, Zhao, Tulyakov, Wu, and Peng}]{ma2021smil}
Ma, M.; Ren, J.; Zhao, L.; Tulyakov, S.; Wu, C.; and Peng, X. 2021.
\newblock Smil: Multimodal learning with severely missing modality.
\newblock In \emph{AAAI}, volume~35, 2302--2310.

\bibitem[{Ma et~al.(2024{\natexlab{b}})Ma, Yang, Li, Hu, Lv, and Peng}]{ma2024cross}
Ma, X.; Yang, M.; Li, Y.; Hu, P.; Lv, J.; and Peng, X. 2024{\natexlab{b}}.
\newblock Cross-modal Retrieval with Noisy Correspondence via Consistency Refining and Mining.
\newblock \emph{TIP}.

\bibitem[{Mummadi et~al.(2021)Mummadi, Hutmacher, Rambach, Levinkov, Brox, and Metzen}]{mummadi2021test}
Mummadi, C.~K.; Hutmacher, R.; Rambach, K.; Levinkov, E.; Brox, T.; and Metzen, J.~H. 2021.
\newblock Test-time adaptation to distribution shift by confidence maximization and input transformation.
\newblock \emph{arXiv preprint arXiv:2106.14999}.

\bibitem[{Nagrani et~al.(2021)Nagrani, Yang, Arnab, Jansen, Schmid, and Sun}]{nagrani2021attention}
Nagrani, A.; Yang, S.; Arnab, A.; Jansen, A.; Schmid, C.; and Sun, C. 2021.
\newblock Attention bottlenecks for multimodal fusion.
\newblock \emph{NeurIPS}, 34: 14200--14213.

\bibitem[{Nguyen et~al.(2023)Nguyen, Nguyen-Tang, Lim, and Torr}]{nguyen2023tipi}
Nguyen, A.~T.; Nguyen-Tang, T.; Lim, S.-N.; and Torr, P.~H. 2023.
\newblock Tipi: Test time adaptation with transformation invariance.
\newblock In \emph{CVPR}, 24162--24171.

\bibitem[{Niu et~al.(2022)Niu, Wu, Zhang, Chen, Zheng, Zhao, and Tan}]{niu2022efficient}
Niu, S.; Wu, J.; Zhang, Y.; Chen, Y.; Zheng, S.; Zhao, P.; and Tan, M. 2022.
\newblock Efficient test-time model adaptation without forgetting.
\newblock In \emph{ICML}, 16888--16905. PMLR.

\bibitem[{Niu et~al.(2023)Niu, Wu, Zhang, Wen, Chen, Zhao, and Tan}]{niu2023towards}
Niu, S.; Wu, J.; Zhang, Y.; Wen, Z.; Chen, Y.; Zhao, P.; and Tan, M. 2023.
\newblock Towards stable test-time adaptation in dynamic wild world.
\newblock \emph{arXiv preprint arXiv:2302.12400}.

\bibitem[{Pei et~al.(2023)Pei, Jiang, Men, Chen, Liu, and Chen}]{pei2023uncertainty}
Pei, J.; Jiang, Z.; Men, A.; Chen, L.; Liu, Y.; and Chen, Q. 2023.
\newblock Uncertainty-induced transferability representation for source-free unsupervised domain adaptation.
\newblock \emph{TIP}, 32: 2033--2048.

\bibitem[{Peng et~al.(2022)Peng, Wei, Deng, Wang, and Hu}]{peng2022balanced}
Peng, X.; Wei, Y.; Deng, A.; Wang, D.; and Hu, D. 2022.
\newblock Balanced multimodal learning via on-the-fly gradient modulation.
\newblock In \emph{CVPR}, 8238--8247.

\bibitem[{Prabhudesai et~al.(2024)Prabhudesai, Ke, Li, Pathak, and Fragkiadaki}]{prabhudesai2024test}
Prabhudesai, M.; Ke, T.-W.; Li, A.; Pathak, D.; and Fragkiadaki, K. 2024.
\newblock Test-time adaptation of discriminative models via diffusion generative feedback.
\newblock \emph{NeurIPS}, 36.

\bibitem[{Prakash, Chitta, and Geiger(2021)}]{prakash2021multi}
Prakash, A.; Chitta, K.; and Geiger, A. 2021.
\newblock Multi-modal fusion transformer for end-to-end autonomous driving.
\newblock In \emph{CVPR}, 7077--7087.

\bibitem[{Shin et~al.(2022)Shin, Tsai, Zhuang, Schulter, Liu, Garg, Kweon, and Yoon}]{shin2022mm}
Shin, I.; Tsai, Y.-H.; Zhuang, B.; Schulter, S.; Liu, B.; Garg, S.; Kweon, I.~S.; and Yoon, K.-J. 2022.
\newblock Mm-tta: multi-modal test-time adaptation for 3d semantic segmentation.
\newblock In \emph{CVPR}, 16928--16937.

\bibitem[{Tan et~al.(2023)Tan, Liu, Long, Jiang, Lu, and Zhang}]{tan2023federated}
Tan, Y.; Liu, Y.; Long, G.; Jiang, J.; Lu, Q.; and Zhang, C. 2023.
\newblock Federated learning on non-iid graphs via structural knowledge sharing.
\newblock In \emph{AAAI}, volume~37, 9953--9961.

\bibitem[{Tang et~al.(2023)Tang, Chen, Niu, Sugiyama, and Gong}]{tang2023distribution}
Tang, J.; Chen, S.; Niu, G.; Sugiyama, M.; and Gong, C. 2023.
\newblock Distribution shift matters for knowledge distillation with webly collected images.
\newblock In \emph{ICCV}, 17470--17480.

\bibitem[{Tang et~al.(2024)Tang, Su, Ye, and Zhu}]{tang2024source}
Tang, S.; Su, W.; Ye, M.; and Zhu, X. 2024.
\newblock Source-Free Domain Adaptation with Frozen Multimodal Foundation Model.
\newblock In \emph{CVPR}, 23711--23720.

\bibitem[{Tsai et~al.(2024)Tsai, Chen, Chen, Yang, Su, Sun, and Kuo}]{tsai2024gda}
Tsai, Y.-Y.; Chen, F.-C.; Chen, A.~Y.; Yang, J.; Su, C.-C.; Sun, M.; and Kuo, C.-H. 2024.
\newblock GDA: Generalized Diffusion for Robust Test-time Adaptation.
\newblock In \emph{CVPR}, 23242--23251.

\bibitem[{Vaswani et~al.(2017)Vaswani, Shazeer, Parmar, Uszkoreit, Jones, Gomez, Kaiser, and Polosukhin}]{vaswani2017attention}
Vaswani, A.; Shazeer, N.; Parmar, N.; Uszkoreit, J.; Jones, L.; Gomez, A.~N.; Kaiser, {\L}.; and Polosukhin, I. 2017.
\newblock Attention is all you need.
\newblock \emph{NeurIPS}, 30.

\bibitem[{Wang et~al.(2021)Wang, Shelhamer, Liu, Olshausen, and Darrell}]{wang2020tent}
Wang, D.; Shelhamer, E.; Liu, S.; Olshausen, B.; and Darrell, T. 2021.
\newblock Tent: Fully Test-Time Adaptation by Entropy Minimization.
\newblock In \emph{ICLR}.

\bibitem[{Wang et~al.(2024)Wang, Luo, Hu, and Zhang}]{wang2024gradient}
Wang, H.; Luo, S.; Hu, G.; and Zhang, J. 2024.
\newblock Gradient-Guided Modality Decoupling for Missing-Modality Robustness.
\newblock In \emph{AAAI}, volume~38, 15483--15491.

\bibitem[{Wang et~al.(2023)Wang, Zhang, Yan, Zhang, and Li}]{wang2023feature}
Wang, S.; Zhang, D.; Yan, Z.; Zhang, J.; and Li, R. 2023.
\newblock Feature alignment and uniformity for test time adaptation.
\newblock In \emph{CVPR}, 20050--20060.

\bibitem[{Woo et~al.(2023)Woo, Lee, Park, Nugroho, and Kim}]{woo2023towards}
Woo, S.; Lee, S.; Park, Y.; Nugroho, M.~A.; and Kim, C. 2023.
\newblock Towards good practices for missing modality robust action recognition.
\newblock In \emph{AAAI}, volume~37, 2776--2784.

\bibitem[{Wu et~al.(2024)Wu, Chi, Wang, Plataniotis, and Feng}]{wu2024test}
Wu, Y.; Chi, Z.; Wang, Y.; Plataniotis, K.~N.; and Feng, S. 2024.
\newblock Test-time domain adaptation by learning domain-aware batch normalization.
\newblock In \emph{AAAI}, volume~38, 15961--15969.

\bibitem[{Xia et~al.(2024)Xia, Huang, Zhu, and Zhao}]{xia2024achieving}
Xia, Y.; Huang, H.; Zhu, J.; and Zhao, Z. 2024.
\newblock Achieving cross modal generalization with multimodal unified representation.
\newblock \emph{NeurIPS}, 36.

\bibitem[{Xiong and Xiang(2024)}]{xiong2024robust}
Xiong, H.; and Xiang, Y. 2024.
\newblock Robust gradient aware and reliable entropy minimization for stable test-time adaptation in dynamic scenarios.
\newblock \emph{The Visual Computer}, 1--16.

\bibitem[{Xu, Yuan, and Ma(2023)}]{xu2023murf}
Xu, H.; Yuan, J.; and Ma, J. 2023.
\newblock Murf: Mutually reinforcing multi-modal image registration and fusion.
\newblock \emph{TPAMI}, 45(10): 12148--12166.

\bibitem[{Xu, Zhu, and Clifton(2023)}]{xu2023multimodal}
Xu, P.; Zhu, X.; and Clifton, D.~A. 2023.
\newblock Multimodal learning with transformers: A survey.
\newblock \emph{TPAMI}, 45(10): 12113--12132.

\bibitem[{Yang et~al.(2022{\natexlab{a}})Yang, Huang, Hu, Li, Lv, and Peng}]{yang2022learning}
Yang, M.; Huang, Z.; Hu, P.; Li, T.; Lv, J.; and Peng, X. 2022{\natexlab{a}}.
\newblock Learning with twin noisy labels for visible-infrared person re-identification.
\newblock In \emph{CVPR}, 14308--14317.

\bibitem[{Yang et~al.(2024)Yang, Li, Zhang, Hu, and Peng}]{yang2024test}
Yang, M.; Li, Y.; Zhang, C.; Hu, P.; and Peng, X. 2024.
\newblock Test-time Adaptation against Multi-modal Reliability Bias.
\newblock In \emph{ICLR}.

\bibitem[{Yang et~al.(2022{\natexlab{b}})Yang, Zhou, Wang, Lu, and Zheng}]{yang2022test}
Yang, T.; Zhou, S.; Wang, Y.; Lu, Y.; and Zheng, N. 2022{\natexlab{b}}.
\newblock Test-time batch normalization.
\newblock \emph{arXiv preprint arXiv:2205.10210}.

\bibitem[{Yao and Wan(2020)}]{yao2020multimodal}
Yao, S.; and Wan, X. 2020.
\newblock Multimodal transformer for multimodal machine translation.
\newblock In \emph{ACL}, 4346--4350.

\bibitem[{Yu et~al.(2021)Yu, Xu, Yuan, and Wu}]{yu2021learning}
Yu, W.; Xu, H.; Yuan, Z.; and Wu, J. 2021.
\newblock Learning modality-specific representations with self-supervised multi-task learning for multimodal sentiment analysis.
\newblock In \emph{AAAI}, volume~35, 10790--10797.

\bibitem[{Zhang et~al.(2023)Zhang, Qi, Shi, and Gao}]{zhang2023domainadaptor}
Zhang, J.; Qi, L.; Shi, Y.; and Gao, Y. 2023.
\newblock Domainadaptor: A novel approach to test-time adaptation.
\newblock In \emph{ICCV}, 18971--18981.

\bibitem[{Zhang et~al.(2022)Zhang, Qiu, Wang, Zeng, Zhang, An, Ma, and Ding}]{zhang2022transformer}
Zhang, W.; Qiu, F.; Wang, S.; Zeng, H.; Zhang, Z.; An, R.; Ma, B.; and Ding, Y. 2022.
\newblock Transformer-based multimodal information fusion for facial expression analysis.
\newblock In \emph{CVPR}, 2428--2437.

\bibitem[{Zhao et~al.(2022)Zhao, Chen, Gao, Wang, Yang, Ren, Xia, and Liu}]{zhao2022target}
Zhao, Y.; Chen, J.; Gao, C.; Wang, W.; Yang, L.; Ren, H.; Xia, H.; and Liu, S. 2022.
\newblock Target-driven structured transformer planner for vision-language navigation.
\newblock In \emph{ACM MM}, 4194--4203.

\bibitem[{Zheng et~al.(2023)Zheng, Li, Chen, Wang, and Luo}]{zheng2023autofed}
Zheng, T.; Li, A.; Chen, Z.; Wang, H.; and Luo, J. 2023.
\newblock Autofed: Heterogeneity-aware federated multimodal learning for robust autonomous driving.
\newblock In \emph{MobiCom}, 1--15.

\bibitem[{Zhou, Chen, and Cao(2020)}]{zhou2020improving}
Zhou, K.; Chen, L.; and Cao, X. 2020.
\newblock Improving multispectral pedestrian detection by addressing modality imbalance problems.
\newblock In \emph{ECCV}, 787--803. Springer.

\bibitem[{Zhu et~al.(2021)Zhu, Xu, Liu, and Jin}]{zhu2021federated}
Zhu, H.; Xu, J.; Liu, S.; and Jin, Y. 2021.
\newblock Federated learning on non-IID data: A survey.
\newblock \emph{Neurocomputing}, 465: 371--390.

\bibitem[{Zong and Sun(2023)}]{zong2023mcomet}
Zong, D.; and Sun, S. 2023.
\newblock Mcomet: Multimodal fusion transformer for physical audiovisual commonsense reasoning.
\newblock In \emph{AAAI}, volume~37, 6621--6629.

\end{thebibliography}

\end{document}